\documentclass[conference]{IEEEtran}

\IEEEoverridecommandlockouts
\usepackage{cite}
\usepackage{amsmath,amssymb,amsfonts}
\usepackage{graphicx}
\usepackage{pifont}
\usepackage{booktabs} 
\usepackage{multirow}
\usepackage{bm}

\usepackage{textcomp}
\usepackage{xcolor}
\def\BibTeX{{\rm B\kern-.05em{\sc i\kern-.025em b}\kern-.08em
    T\kern-.1667em\lower.7ex\hbox{E}\kern-.125emX}}
\usepackage{url}  
\usepackage{hyperref}
\usepackage{rotating}
\usepackage{xcolor}
\usepackage{array}
\usepackage{threeparttable}
\usepackage{titlesec}
\usepackage{siunitx}

\newcolumntype{C}[1]{>{\centering\arraybackslash}p{#1}}

\begin{document}

\title{SoK: Membership Inference Attacks on LLMs are Rushing Nowhere (and How to Fix It)\thanks{This work has been accepted for publication in the IEEE Conference on Secure and Trustworthy Machine Learning (SaTML, 2025). The final version will be available on IEEE Xplore.

The code and data required to replicate the results in this paper can be found here: \url{https://github.com/computationalprivacy/mia_llms_benchmark}.
}
}

\author{\textit{Anonymous authors}}

\author{
    Matthieu Meeus\textsuperscript{1},
    Igor Shilov\textsuperscript{1},
    Shubham Jain\textsuperscript{2},
    Manuel Faysse\textsuperscript{3},
    Marek Rei\textsuperscript{1},
    Yves-Alexandre de Montjoye\textsuperscript{1}
    \\[10pt]
    \textsuperscript{1}\textit{Imperial College London}\\
    \textsuperscript{2}\textit{Sense Street}\\
    \textsuperscript{3}\textit{MICS, CentraleSupélec, Université Paris-Saclay}
}

\date{}

\maketitle

\begin{abstract}
Whether Large Language models (LLMs) memorize their training data and what this means, from measuring privacy leakage to detecting copyright violations, has become a rapidly growing area of research. In the last few months, more than 10 new methods have been proposed to perform Membership Inference Attacks (MIAs) against LLMs. Contrary to traditional MIAs which rely on fixed---but randomized---records or models, these methods are mostly trained and tested on datasets collected post-hoc. Sets of members and non-members, used to evaluate the MIA, are constructed using informed guesses after the release of a model. This lack of randomization raises concerns of a distribution shift between members and non-members. In this work, we first extensively review the literature on MIAs against LLMs and show that, while most work focuses on sequence-level MIAs evaluated in post-hoc setups, a range of target models, motivations and units of interest are considered. We then quantify distribution shifts present in 6 datasets used in the literature, ranging from books to papers using a model-less bag of word classifier and compare the model-less results to the MIA. Our analysis shows that all of these datasets constructed post-hoc suffer from strong distribution shifts. These shifts invalidate the claims of LLMs memorizing strongly in real-world scenarios and, potentially, also the methodological contributions of the recent papers based on these datasets. Yet, all hope might not be lost. In the second part of this work, we introduce important considerations to properly evaluate MIAs against LLMs and discuss, in turn, potential ways forwards: randomized test splits, injections of randomized (unique) sequences, randomized fine-tuning, and several post-hoc control methods. While each option comes with its advantages and limitations, we believe they collectively provide solid grounds to guide the development of MIA methods and study LLM memorization. We conclude with an overview of recommended approaches to benchmark sequence-level and document-level MIAs against LLMs.
\end{abstract}

\section{Introduction}
The prevalence of Large Language Models (LLMs)~\cite{gpt4techreport,llama3modelcard,jiang2023mistral} has sparked significant interest in studying LLM memorization to better understand how they acquire their capabilities and learn facts. LLMs can also be trained on confidential or sensitive personal data, making the study of LLM memorization critical from both an information security and a privacy perspective~\cite{brown2022does,mireshghallah2022empirical,carlini2019secret,lukas2023analyzing,carlini2021extracting,mireshghallah2023can}. Similarly, research helps assess whether LLMs are being trained on copyrighted material~\cite{shi2023detecting,wei2024proving,meeuscopyright,duarte2024cop} and the extent to which they might generate such content~\cite{karamolegkou2023copyright}. Further, to fairly evaluate newly released models, it is crucial to understand whether they are trained on (and memorize) evaluation benchmarks~\cite{magar2022data,li2023estimating,oren2023proving,golchin2023data,deng2023investigating,li2024task,dong2024generalization}. 

Membership Inference Attacks (MIAs) have become a widely used tool to study the memorization of machine learning (ML) models~\cite{shokri2017membership,carlini2022membership}. In an MIA, an attacker aims to infer whether a certain data sample has been used to train the target model. Recently, MIAs have also been proposed for LLMs~\cite{carlini2021extracting,yeom2018privacy,mattern2023membership,shi2023detecting,meeuscopyright,zhang2024min,xie2024recall,liu2024probing,ye2024data,zhang2024adaptive,wang2024recall,mozaffari2024semantic,kaneko2024sampling} to infer whether a specific piece of text has been used to train an LLM. While most of these works have focused on sequences as the unit of interest to infer membership (typically $\leq 256$ tokens), other works have considered document~\cite{meeus2024did,shi2023detecting}, and dataset-level MIAs~\cite{maini2024llm}. Beyond MIAs, prior work has also widely studied training data extraction from LLMs, either verbatim~\cite{carlini2022quantifying,nasr2023scalable,zhang2023make} or not~\cite{ippolito2022preventing,peng2023near,karamolegkou2023copyright}, to measure LLM memorization. 

Independently of how MIAs against ML models (e.g. image classification) are developed,  they are traditionally \emph{evaluated} in randomized experiments~\cite{shokri2017membership,carlini2022membership}, instantiated as a privacy game between a model developer and an attacker. Ye et al.~\cite{ye2022enhanced} proposes to distinguish between privacy games with \emph{fixed records} and \emph{fixed models}. For fixed records, multiple models are trained with the target randomly included or excluded~\cite{long2018understanding,long2020pragmatic,stadler2022synthetic,jagielski2020auditing}. Fixed model scenarios use a single model trained on randomly sampled, IID records~\cite{nasr2019comprehensive,salem2018ml,watson2021importance}. While feasible for traditional ML models, evaluating MIAs using a privacy game for LLMs presents unique challenges. On the one hand, the mere scale and cost of training indeed prohibits the development of LLMs solely for the purpose of playing a privacy game. On the other hand, it remains uncertain how findings from small-scale MIA experiments translate to the scale of real-world LLMs.

As a result, recent MIAs against LLMs have been evaluated using observational data collected \emph{post-hoc}. Two concurrent works, published at ICLR~\cite{shi2023detecting} and USENIX Security~\cite{meeus2024did} in 2024, first instantiate the membership inference task post-hoc; where an attacker aims to apply an MIA on a released LLM without access to the randomized training data distribution. Authors both use publicly available datasets widely used for LLM training to collect likely members, and use data released on the same sources but after the LLM's training data cutoff date to collect likely non-members. Both contributions report MIAs to work very well, with AUCs up to $0.8$. Since then, numerous pieces of work used either the same datasets of members and non-members, or similar datasets collected post-hoc, to develop and evaluate MIAs against LLMs~\cite{zhang2024min,xie2024recall,liu2024probing,ye2024data,zhang2024adaptive,wang2024recall,mozaffari2024semantic,kaneko2024sampling}, all reporting high AUCs. 

In parallel with new MIAs being developed, strong concerns have been raised on the lack of randomization in post-hoc collection of members and non-members. Duan et al.~\cite{duan2024membership} first suggested that this setup introduces a distribution shift between members and non-members due to the temporal gap in data collection. These concerns were shared by Maini et al.~\cite{maini2024llm} and later also by Das et al.~\cite{das2024blind}\footnote{Das et al.~\cite{das2024blind} is concurrent with this work (for a comparison see Sec.~\ref{sec:compare_concurrent}).}. Such distribution shifts would lead membership classifiers to learn temporal differences in language rather than membership, potentially inflating AUC scores and leading to an incorrect understanding of LLM memorization. 

\textbf{In this work, we provide an overview of the recent surge in research on MIAs against LLMs, identifying numerous works using datasets collected post-hoc used for MIA evaluation. With a bag of words classifier, we then identify serious distribution shifts present in these datasets. Lastly, we carefully examine the requirements for rigorous MIA evaluation and solutions moving forward.}

First, we formalize the task of MIAs against LLMs, distinguishing between MIAs targeting LLM pretraining and finetuning, on the sequence-, document- and dataset-level. In Table~\ref{tab:overview}, we review the literature, with a focus on the most recent advances ('23-'24) while including seminal initial work. We find that only in '24, at least $10$ new MIA methods have been proposed. Next, we find that the primary motivation for such studies has evolved over time, with initial work mostly studying LLM finetuning for privacy and security reasons~\cite{carlini2019secret,carlini2021extracting,mireshghallah2022empirical,mattern2023membership,lukas2023analyzing}, while more recent work studies MIAs to identify copyright violations or data contamination, increasingly considering LLM pretraining. We then carefully examine which datasets have been considered to evaluate MIAs across the literature. We find that the majority of recent work does not consider randomized privacy games, and instead relies on data collected post-hoc (flagged with \textcolor{red}{\checkmark} in Table~\ref{tab:overview}). 

Second, we introduce a \emph{bag of words} classifier to examine the distribution shift between members and non-members as flagged by prior~\cite{duan2024membership,maini2024llm} and concurrent work~\cite{das2024blind}. We train the bag of words classifier to distinguish members from non-members, without considering the target model and use its performance to quantify the distribution shift and understand its impact on previously reported results. We apply this to $6$ datasets collected post-hoc~\cite{meeus2024did,shi2023detecting,li2022auditing,ye2024data}  (used in over $10$ papers for MIA evaluation) and confirm that the simple classifier yields an AUC often better than proposed MIAs. This shows that there exists such a serious distribution shift in language between the collected members and non-members that it is impossible to exclusively attribute the MIA performance to LLM memorization. This questions the effectiveness of newly proposed MIAs~\cite{zhang2024min,xie2024recall,liu2024probing,ye2024data,zhang2024adaptive,wang2024recall,mozaffari2024semantic,kaneko2024sampling} (latest one from September, 2024), potentially leading to false conclusions on LLM memorization.

Third, before examining the potential solutions to evaluate MIAs against LLMs, we elaborate on two core challenges. First, the inherent overlap between texts collected from the internet challenges the binary notion of membership. Prior work found significant overlap between randomized splits from the training data, either trying to achieve a clean model validation set~\cite{brown2020language,wei2021finetuned,du2022glam,touvron2023llama2} or a useful set of members and non-members to evaluate MIAs~\cite{duan2024membership}. Second, we elaborate on the unit of interest considered in MIAs against LLMs, ranging from a sequence to an entire dataset. 

Lastly, we examine the approaches available for rigorous evaluation of MIAs against LLMs in the future. We consider the following $4$ options and elaborate on their advantages and limitations:

\begin{enumerate}
    \item The use of the randomized split of train and test data released for open-source LLMs as members and non-members, as suggested by prior work~\cite{li2023mope,duan2024membership,maini2024llm}.
    \item The injection of randomized unique sequences in LLM training data, as previously used to study LLM memorization~\cite{henderson2018ethical,carlini2019secret,thakkar2020understanding,thomas2020investigating,wei2024proving,meeuscopyright}. 
    \item The creation of a randomized evaluation setup by finetuning LLMs on smaller datasets, as proposed before~\cite{mattern2023membership,mireshghallah2022empirical,zeng2023exploring,lee2023language,lukas2023analyzing}. 
    \item The application of control methods to make evaluation and calibration of MIAs against LLMs feasible on data collected post-hoc, either with a skewed prior~\cite{jayaraman2020revisiting,kazmi2024panoramia} or de-biasing techniques~\cite{eichler2024nob} as proposed by recent work, or as we here propose using techniques inspired by Regression Discontinuity Design (RDD)~\cite{lee2010regression}.
\end{enumerate}

We conclude with an overview of recommended approaches to benchmark newly proposed sequence and document-level MIAs in the future (Table~\ref{tab:benchmark_proposal}). We hope this overview contributes to a better understanding of the recent advances in MIAs against LLMs and provides a solid foundation for future research to rigorously evaluate and interpret MIA results.

\begin{table*}[ht]
\centering
\scriptsize
\caption{Overview of prior work studying MIAs against LLMs. We distinguish target models, tasks, motivations and datasets.}
\begin{threeparttable}
\setlength{\tabcolsep}{4pt}  
\begin{tabular}{l|cc|cc|c|ccc|ccccccc}
\toprule
&
\multicolumn{2}{c|}{Target} & 
\multicolumn{2}{c|}{Unit of interest} &
& 
\multicolumn{3}{c|}{Motivation} & 
\multicolumn{7}{c|}{Dataset} 
\\
\midrule
MIA study & 
\rotatebox[origin=c]{90}{\shortstack{Finetuning}} &
\rotatebox[origin=c]{90}{\shortstack{Pretraining}} &
\rotatebox[origin=c]{90}{Sequence} &
\rotatebox[origin=c]{90}{\shortstack{Document \\ or dataset}} &
\rotatebox[origin=c]{90}{\shortstack{New MIA \\ method}} & 
\rotatebox[origin=c]{90}{\shortstack{Privacy \& \\ Security}} & 
\rotatebox[origin=c]{90}{Copyright} & 
\rotatebox[origin=c]{90}{\shortstack{Dataset \\ contamination}} & 
\rotatebox[origin=c]{90}{\textcolor{red}{WikiMIA\tnote{*}~\cite{shi2023detecting}}} &
\rotatebox[origin=c]{90}{\textcolor{red}{Arxiv\tnote{*}}} &
\rotatebox[origin=c]{90}{\textcolor{red}{Books\tnote{*}}} &
\rotatebox[origin=c]{90}{\textcolor{red}{\shortstack{Stack \\ Exchange\tnote{*}}}} &
\rotatebox[origin=c]{90}{Pile~\cite{pile}} &
\rotatebox[origin=c]{90}{MIMIR\tnote{1}~\cite{duan2024membership}} &
\rotatebox[origin=c]{90}{\shortstack{Randomized \\ sequence}} 
\\

\midrule
\midrule
('19) Carlini et al.~\cite{carlini2019secret} 
& \checkmark &
& \checkmark & 
& \checkmark
& \checkmark & & 
& & & & & & & \checkmark 
 \\
\midrule
('21) Carlini et al.~\cite{carlini2021extracting} 
& \checkmark &
& \checkmark &
& \checkmark
& \checkmark & & 
& & & & & & & 
\\
\midrule
('22) Mireshghallah et al.~\cite{mireshghallah2022empirical} 
& \checkmark &
& \checkmark &
&
& \checkmark & &
& & & & & & & \checkmark 
\\
\midrule
('23) Mattern et al.~\cite{mattern2023membership}
& \checkmark &
& \checkmark &
& \checkmark 
& \checkmark & & 
& & & & & & & \checkmark 
\\
\midrule
('23) Lukas et al.~\cite{lukas2023analyzing} 
& \checkmark &
& \checkmark & 
& 
& \checkmark & & 
& & & & & & & \checkmark \\
\midrule
('23) Shi et al.~\cite{shi2023detecting} 
&  & \checkmark
& \checkmark & \checkmark 
& \checkmark
& & \checkmark & \checkmark 
& \textcolor{red}{\checkmark} & & \textcolor{red}{\checkmark} & & & & 
\\
\midrule
('23) Meeus et al.~\cite{meeus2024did} 
& & \checkmark
& & \checkmark 
& \checkmark 
& & \checkmark & \checkmark 
& & \textcolor{red}{\checkmark} & \textcolor{red}{\checkmark} & & & & 
\\
\midrule
('23) Li et al.~\cite{li2023mope} 
& & \checkmark
& \checkmark &
& \checkmark
& \checkmark &  & 
& & & & & \checkmark & & 
\\
\midrule
('24) Duan et al.~\cite{duan2024membership} 
& & \checkmark
& \checkmark &
& 
& \checkmark & \checkmark & \checkmark 
& & & & & & \checkmark &  
\\
\midrule
('24) Zhang et al.~\cite{zhang2024min} 
& & \checkmark
& \checkmark &
& \checkmark 
& & \checkmark & \checkmark 
& \textcolor{red}{\checkmark} & & & & & \checkmark & \\
\midrule
('24) Xie et al.~\cite{xie2024recall} 
& & \checkmark
& \checkmark &
& \checkmark 
& & \checkmark & \checkmark 
& \textcolor{red}{\checkmark} & & & & & \checkmark & \\
\midrule
('24) Liu et al.~\cite{liu2024probing} 
& & \checkmark
& \checkmark &
& \checkmark 
& & \checkmark & \checkmark 
& \textcolor{red}{\checkmark} & \textcolor{red}{\checkmark} & & & & & \\
\midrule
('24) Ye et al.~\cite{ye2024data} 
& & \checkmark
& \checkmark &
& \checkmark 
& & & \checkmark
& \textcolor{red}{\checkmark} & & & \textcolor{red}{\checkmark} & & & \\
\midrule
('24) Zhang et al.~\cite{zhang2024adaptive} 
& & \checkmark
& \checkmark & 
& \checkmark 
& \checkmark & \checkmark & \checkmark 
& \textcolor{red}{\checkmark} & & \textcolor{red}{\checkmark} & & & \checkmark &  \\
\midrule
('24) Wang et al.~\cite{wang2024recall} 
& & \checkmark
& \checkmark &
& \checkmark 
& \checkmark & \checkmark & \checkmark 
& \textcolor{red}{\checkmark} & & & & & \checkmark & \\
\midrule
('24) Mozaffari et al.~\cite{mozaffari2024semantic} 
& & \checkmark
& \checkmark &
& \checkmark 
& \checkmark & \checkmark & \checkmark
& \textcolor{red}{\checkmark\tnote{2}} & & & & & \checkmark & \\
\midrule
('24) Kaneko et al.~\cite{kaneko2024sampling} 
& & \checkmark
& \checkmark &
& \checkmark 
& \checkmark & \checkmark & \checkmark 
& \textcolor{red}{\checkmark} & & & & & & \\
\midrule
('24) Duarte et al.~\cite{duarte2024cop} 
& & \checkmark
& \checkmark & 
& \checkmark 
&  & \checkmark &  
& & & \textcolor{red}{\checkmark}& & & & \\
\midrule
('24) Miani et al.~\cite{maini2024llm} 
& & \checkmark
& & \checkmark 
& \checkmark
& \checkmark & \checkmark & \checkmark &
&  & & & & \checkmark &  \\
\midrule
('24) Wei et al.~\cite{wei2024proving} 
& \checkmark & \checkmark 
& \checkmark &
& 
& & \checkmark & \checkmark
& & & & \checkmark\tnote{3} & & & \checkmark \\
\midrule
('24) Meeus et al.~\cite{meeuscopyright} 
& & \checkmark
& \checkmark & 
& 
& & \checkmark & 
& & & & & & & \checkmark\\
\midrule
('24) Panaitescu et al.~\cite{panaitescu2024can} 
& & \checkmark
& \checkmark & 
& 
& & \checkmark & 
& \textcolor{red}{\checkmark} & & & & & & \\
\midrule
\bottomrule
\end{tabular}
\begin{tablenotes}
  \scriptsize
  \item[*] Red indicates likely presence of the distribution shift
  \item[1] While MIMIR originates from a randomized setup, subsquent deduplication~\cite{duan2024membership} could introduce a distribution shift (Sec.~\ref{sec:pythia_solution}).
  \item[2] Custom datasets with the same underlying data as the indicated dataset.
  \item[3] SHA-hashes extracted from the StackExchange datasets, likely does not suffer from the distribution shift.
\end{tablenotes}
\label{tab:overview}
\end{threeparttable}
\end{table*}

\section{Preliminary}
\label{sec:preliminary}

\textbf{Threat model.} We consider a target LLM $\mathcal{M}$ trained on dataset $\mathcal{D}$, with tokenizer $T$. The goal of the attacker $\mathcal{A}$ is to leverage access to the LLM $\mathcal{M}$ to infer whether a specific piece of text has been part of the training data $\mathcal{D}$. 

MIAs against LLMs have been proposed across varying assumptions made for the attacker $\mathcal{A}$, from access to model weights (white-box)~\cite{li2023mope,liu2024probing}, to only query access with model logits (black-box)~\cite{yeom2018privacy,carlini2021extracting,mattern2023membership,shi2023detecting,meeus2024did,zhang2024min,xie2024recall,ye2024data,zhang2024adaptive,wang2024recall,mozaffari2024semantic} or merely generated text~\cite{duarte2024cop,kaneko2024sampling} (no-box). Some work also assumes access to additional non-member data from the same distribution as the target sample~\cite{xie2024recall,liu2024probing,wang2024recall}. We here focus on the evaluation of MIAs against LLMs, independent of the threat model used.

\textbf{Dataset Structure for LLM Training.} We formalize the LLM training dataset $\mathcal{D}$ as a collection of $N$ documents: $\mathcal{D} = \{d_1, d_2, \ldots, d_N\}$, where each document $d_i$ is composed of a sequence of tokens. A document is an ordered sequence of tokens, $d_i = (t_1, t_2, ..., t_{N_{d_i}})$, with $N_{d_i}$ the length of document $d_i$. We consider documents originating from a single, identifiable source which are coherent and semantically meaningful when considered in isolation. Examples include a news article, blogpost, book, academic paper or piece of code. The document length can vary widely and is a property of the document itself, thus independent of whether the entire document fits into the LLM's context window. For a single book this can be up to more than 100k tokens, while for social media posts this can be under $100$ tokens.  

Given a document $d_i = (t_1, t_2, ..., t_{N_{d_i}})$, a sequence $s_{ij}$ is  a contiguous subsequence of tokens from $d_i$. Formally, for some starting index $a$ and ending index $b$, where $1 \leq a \leq b \leq N_{d_i}$, a sequence $s_{ij}$ is defined as $s_{ij} = (t_a, t_{a+1}, ..., t_b)$. The set of all possible sequences from document $d_i$ is denoted as $\mathcal{S}_i$, where $\mathcal{S}_i = \{(t_a, t_{a+1}, ..., t_b) \mid 1 \leq a \leq b \leq N_{d_i}\}$. The length of a sequence $s_{ij}$, denoted as $|s_{ij}|$, is defined as $b - a + 1$. Sequences are not required to align with semantic boundaries in the document (e.g., sentences, paragraphs) and are defined by their start and end indices within the document. The sequence length can be up to the entire document ($a = 1$ and $b = N_{d_i}$), although sequences are typically assumed to be shorter. The choice of sequence boundaries $(a, b)$ have been considered random (random excerpts from books)~\cite{shi2023detecting} or based on specific criteria (e.g. start of a document~\cite{duan2024membership}).

\textbf{LLM training.} Tokenizer $T$ maps a piece of text $S_{ij}$ to a sequence of tokens $T(S_{ij}) = s_{ij} = (t_a, t_{a+1}, \ldots, t_b)$, where each token belongs to a vocabulary $\mathcal{V}$ of size $|\mathcal{V}| = V$. Given a sequence of tokens (referred to as the context), auto-regressive language models predict the probability distribution for the next token over all tokens in vocabulary $\mathcal{V}$. This capability is acquired during training on the entire dataset $\mathcal{D}$, where the model minimizes the loss, generally cross-entropy loss, for next-token prediction.
For a model $\mathcal{M}$ with parameters $\theta$, we define the loss computed on sequence $s_{ij}$ as $\mathcal{L}_{\mathcal{M}}(s_{ij}) = -\frac{1}{b-a+1}\sum_{k=a}^{b} \log\left( \mathcal{M}_{\theta}(t_k | t_a, \ldots, t_{k-1})\right)$, where $\mathcal{M}_{\theta}(t_k | t_a, \ldots, t_{k-1})$ represents the probability predicted by the model for the true token $t_k$ given the preceding tokens.

In this work, we exclusively consider models trained on causal language modeling objectives. This includes both base model versions---trained on large quantities of unsupervised data---and versions that have been further fine-tuned on supervised task-specific data to unlock further abilities (instruction following, chat, specific tasks). In Table~\ref{tab:overview}, we refer to these models as pretrained and finetuned, respectively. In this work, we do not consider models trained with more complex alignment techniques such as reinforcement learning methods.

\textbf{Sequence-level MIAs.} The unit of interest in MIAs against LLMs has most commonly been a sequence. A sequence-level MIA aims to determine whether a given sequence $s_t$ has been used to train target LLM $\mathcal{M}$. Given an LLM $\mathcal{M}$ trained on dataset $\mathcal{D}$ and a target sequence $s_t$, a sequence-level MIA is a scoring function $f_{\text{seq}}$ such that $
f_{\text{seq}}(\mathcal{M}, s_t)$ outputs a continuous value representing the likelihood that $s_t$ was included in $\mathcal{D}$. A higher score indicates a higher likelihood of membership. To make a binary prediction $\hat{y}$, a threshold $\tau$ can be applied $\hat{y}_{\text{seq}}(\mathcal{M}, s_t) = 1 \text{ if } f_{\text{seq}}(\mathcal{M}, s_t) \geq \tau_{\text{seq}} \text{, and } 0 \text{ otherwise}$. The MIA performance is then determined with the true label for membership $y$, $y_{\text{seq}}(\mathcal{M}, s_t) = 1 \text{ if } \exists d_i \in \mathcal{D}, s_t \in d_i \text{, and } 0 \text{ otherwise}$. MIAs are typically evaluated on the scoring function $f_{\text{seq}}$ using classification metrics that do not rely on a particular threshold, such as area under the receiver operating characteristic curve (AUC) or true positive rate (TPR) at low false positive rate (FPR)~\cite{carlini2022membership}. 

Most prior studies on MIAs against LLMs have considered sequences of a fixed, often truncated length significantly shorter than most documents. Carlini et al.~\cite{carlini2021extracting} use sequences of $256$ tokens, while Shi et al.~\cite{shi2023detecting} and subsequent works~\cite{zhang2024min,xie2024recall,liu2024probing,ye2024data,zhang2024adaptive,wang2024recall,mozaffari2024semantic,kaneko2024sampling} consider random excerpts between $32$ and $256$ tokens and Meeus et al.~\cite{meeuscopyright} considers up to $100$ synthetically generated tokens. We thus primarily focus on sequence-level MIAs and discuss other units of interest (i.e. documents and datasets, typically longer than sequences) in Sec.~\ref{sec:unit_of_interest} and~\ref{sec:solution_mias}.  

An MIA against an LLM thus involves designing the scoring function $f_{\text{seq}}(\mathcal{M}, s_t)$ which assesses the likelihood that a sequence $s_t$ was observed by model $\mathcal{M}$. For instance, model loss can be used as a naive scoring function (i.e. $f_{\text{seq}}(\mathcal{M}, s_t) = -\mathcal{L}_{\mathcal{M}}(s_t)$)~\cite{yeom2018privacy}, as model training minimizes entropy values for sequences included in the training data. Individual sequences are, however, not seen in isolation during training, as they are likely conditioned on preceding context. Especially for sequences smaller than documents (or data grouped during training), this might impact the effectiveness of using the model loss as a scoring function in sequence-level MIAs against LLMs compared to other ML models. MIAs against LLMs have thus proposed more advanced scoring functions, by, for instance, normalizing the model loss with respect to how 'normal' a piece of text is~\cite{carlini2021extracting}, contrasting the model loss with the loss of neighboring samples~\cite{mattern2023membership} or by only focusing on the lowest predicted probabilities~\cite{shi2023detecting}.

\section{Literature review}

In Table~\ref{tab:overview}, we review the literature studying MIAs against LLMs. We focus on the most recent work ('23-'24) while also including initial seminal work. 

First, we distinguish between MIAs applied to finetuned and pretrained models, as defined in Sec.~\ref{sec:preliminary}. We find that most recent MIAs are primarily targeting pretrained LLMs rather than models finetuned on smaller datasets. 

Second, we distinguish between MIAs based on the unit of interest they consider. Most commonly, prior work has considered sequence-level MIAs, while some exceptions include scaling MIAs to the level of a document~\cite{shi2023detecting,meeus2024did} or dataset~\cite{maini2024llm}. In the rest of this work we focus primarily on sequence-level MIAs, consistent with the majority of the literature, and will revisit document-level MIAs below (Sec.~\ref{sec:unit_of_interest}).

Third, we distinguish between studies developing new MIA methodologies (i.e. designing novel scoring functions $f_{\text{seq}}(\mathcal{M}, s_t)$) and studies merely evaluating prior MIA methodologies in specific setups. In Table~\ref{tab:overview} we show that there is a significant body of recent work focused on developing new MIAs, often claiming to more effectively capture the likelihood of membership than prior work~\cite{carlini2021extracting,yeom2018privacy,mattern2023membership,shi2023detecting,zhang2024min,zhang2024min,xie2024recall,liu2024probing,ye2024data,zhang2024adaptive,wang2024recall,mozaffari2024semantic,kaneko2024sampling,li2023mope}. Studies not proposing new MIAs still evaluate MIAs on particular setups, e.g. to evaluate the effectiveness of privacy-preserving technologies~\cite{chhun2024language,panaitescu2024can}, detect the use of copyrighted content~\cite{meeuscopyright,wei2024proving} or draw connections between MIAs and PII leakage~\cite{lukas2023analyzing}. 

Fourth, we distinguish the primary motivation of each MIA study as either privacy and security, copyright or dataset contamination. While initial work was primarily studying MIAs from a privacy and security perspective~\cite{carlini2019secret,carlini2021extracting,mireshghallah2022empirical,mattern2023membership,lukas2023analyzing}, we find that recent work increasingly considers novel questions associated with LLM pretraining, for instance detecting the inclusion of copyright-protected content or evaluation benchmarks. 

Finally, we also classify the datasets which have been used to study MIAs. Importantly, we identify that most of the recent studies have relied on datasets collected post-hoc for MIA evaluation (\textcolor{red}{\checkmark}). 

\section{Current evaluation methods are flawed}
\label{sec:flawed}

In this section, we introduce a bag of words classifier as a model-less baseline to distinguish members from non-members and use this to examine the validity of using datasets collected post-hoc for MIA evaluation, as employed by most recent work (\textcolor{red}{\checkmark} in Table~\ref{tab:overview}). 

\subsection{MIAs relying on post-hoc data collection}

The evaluation of MIAs is typically instantiated as a randomized privacy game~\cite{ye2022enhanced}. However, creating such a privacy game becomes challenging for LLMs, as the mere scale and cost of training prohibits the development of several LLMs (\emph{fixed target} in~\cite{ye2022enhanced}) or even one (\emph{fixed model} in~\cite{ye2022enhanced}) solely for the purpose of evaluating MIAs. 

Two concurrent works~\cite{shi2023detecting,meeus2024did} first developed strategies to create a set of likely members and likely non-members using information about training data sources and the LLM release date. We refer to such data collection strategies as \emph{post-hoc}, where an attacker aims to infer membership for a released LLM, without access to the randomized training data distribution. In both works, members are collected by sampling documents from the often publicly available datasets widely used for LLM training (e.g. Wikipedia articles~\cite{shi2023detecting}, books in the public domain, ArXiv papers~\cite{meeus2024did}). Non-members are collected as documents from the same data sources but released after the LLM's training data cutoff date. Using this collection technique for evaluation, Shi et al.~\cite{shi2023detecting} propose a novel MIA scoring function based on the lowest predicted probabilities returned by the LLM. Authors achieved a membership AUC of $0.74$ for LLaMA-30B~\cite{touvron2023llama} on the WikiMIA dataset, and an AUC of $0.88$ for GPT-3~\cite{brown2020language} on copyright-protected books (BookMIA). Meeus et al.~\cite{meeus2024did} develop a custom membership classifier against OpenLLaMA~\cite{openlm2023openllama} and achieved MIA AUC of $0.86$ and $0.68$ for books and academic papers, respectively. 

Such post-hoc approaches to collect datasets for MIA evaluation are not randomized, raising the question whether they sufficiently approximate randomness in a privacy game and if biases may be introduced during the data collection process. Indeed, doubts have recently been raised on whether this construction of members and non-members constitutes a distribution shift due to the temporal gap in data collection~\cite{duan2024membership,maini2024llm}, even if efforts to mitigate any distribution shift are employed (e.g. Meeus et al.~\cite{meeus2024did} ensured that the original publication dates of member and non-member books were similarly distributed). If a distribution shift were to be present, the membership classifier might learn the temporal gap in language rather than membership, potentially questioning the validity of MIA results obtained. 

\subsection{Quantifying a potential distribution shift}
\label{sec:distribution_shift}

Duan et al.~\cite{duan2024membership} was the first to flag that the previously reported high MIA AUC might be due to the distribution shift between members and non-members. They observe a significant change in $7$-gram overlap between members and non-members, and also show how the MIA AUC increases when the temporal gap between members and non-members increases. Shortly afterwards, Maini et al.~\cite{maini2024llm} also stated that post-hoc data collection may lead to a confounding variable, making members and non-members not IID. 

We here introduce a bag of words (BoW) classifier to distinguish members from non-members, without taking the target model from the MIA into account. The performance of this classifier reflects what is achievable purely based on the documents, and thus acts as a model-less baseline for the MIA. If this classifier outperforms random guessing in the binary classification task based solely on the dataset’s vocabulary, it may indicate a language shift. 

We use a random forest classifier ($500$ trees with maximum depth of $2$ and minimum samples per leaf of $10$) with the count of words ($1$-grams) appearing in at least $5\%$ of the training documents as features. For each dataset, we train the classifier on $80\%$ of all documents, and evaluate it on the remaining $20\%$ (averaged over $5$ runs, with balanced membership). To understand the classifier's performance, we also identify the words contributing most significantly to the classifier’s decisions.

We instantiate the bag of words classifier on all (available) datasets collected post-hoc and used to evaluate MIAs against LLMs. The WikiMIA and BookMIA datasets, introduced by Shi et al.~\cite{shi2023detecting}, contain excerpts (of $32$ and $128$ tokens) from Wikipedia `events' and copyright protected books before and after early 2023. Gutenberg (full), introduced by Meeus et al.~\cite{meeus2024did} contains full books released on Project Gutenberg before and after the release date of OpenLLaMA~\cite{openlm2023openllama}, while Gutenberg (mid) contains the same books but with the first and last 10\% removed. We also use ArXiv from Meeus et al.~\cite{meeus2024did}, containing academic papers before and after the release date of OpenLLaMA~\cite{openlm2023openllama}. Similarly, ArXivMIA as introduced by Liu et al.~\cite{liu2024probing} contains paper abstracts before and after early 2024. Lastly, we consider StackMIA, as introduced by Ye et al.~\cite{ye2024data}, which contains samples from Stack Exchange before and after early 2023. 

\begin{table}[t]
    \centering
    \caption{AUC (mean$\pm$standard deviation) for the bag of words (BoW) classifier across datasets collected post-hoc.}
    \begin{tabular}{ccc}
    \toprule
        Dataset & BoW AUC & Predictive words (sample)\\
        \midrule 
        \midrule
        WikiMIA (32)~\cite{shi2023detecting}& $0.988 \pm .004$ & \{{2014,2023,2022}\} \\ 
        \midrule
        WikiMIA (128)~\cite{shi2023detecting}& $0.994 \pm .003$ & \{{2014,2023,2022, was}\} \\ 
        \midrule
        BookMIA~\cite{shi2023detecting} & $0.937 \pm .004$ & \{{didn,wasn,upon,hadn}\} \\
        \midrule
        Gutenberg (full)~\cite{meeus2024did} & $0.997 \pm .002$ & \{gutenberg,ebook,online,http\} \\ 
         \midrule
        Gutenberg (mid)~\cite{meeus2024did} & $0.889 \pm .021$ & \{colored,n****,mediæval\} \\ 
         \midrule
         ArXiv~\cite{meeus2024did} & $0.690 \pm .024$ & \{mathcal,2007,2010,prove\} \\ 
        \midrule
        ArXivMIA~\cite{liu2024probing} & $0.596 \pm .037$ & \{training,mathbb,github\} \\ 
         \midrule
        StackMIAsub~\cite{ye2024data} & $0.612 \pm .004$ & \{magnetic,people,surface\} \\ 
        \midrule 
        \bottomrule \\
    \end{tabular}
    \label{tab:bag_of_words}
\end{table}

Table~\ref{tab:bag_of_words} shows that the bag of words classifier can distinguish members from non-members significantly better than a random guess (AUC of $0.5$). For some datasets, the classifier reaches near-perfect AUC, i.e. $0.99$ for the WikiMIA dataset, $0.94$ for the BookMIA dataset and $0.97$ for the books from Project Gutenberg (full), strikingly exceeding the MIA AUC reported by attacks utilizing the corresponding target model~\cite{shi2023detecting,meeus2024did}. For other datasets, the performance is relatively lower, yet remains significantly better (AUC of $\geq0.6$) than a random guess baseline. \textbf{This means that, all datasets relying on post-hoc data collection which have been released, suffer from a significant and often strong distribution shift}. Moreover,  the simplicity of the bag of words classifier suggests that it likely provides a lower bound on classification performance. More advanced classifiers could potentially detect other, subtler distribution shifts.

To further understand the effectiveness of the classifier and the underlying biases in the collected data, we examine the most predictive words for each dataset. 

For WikiMIA~\cite{shi2023detecting}, most predictive words correspond to years. As the dataset is constructed using text from Wikipedia `events', entries almost always contain dates. Data collected for different time periods could thus be distinguished using the event date mentioned in the excerpts. This is precisely what the bag of words classifier exploits, as its top predictive words are primarily years. Similarly for ArXiv papers used by Meeus et al.~\cite{meeus2024did}, years (in citations) carry the most predictive value. For ArXivMIA~\cite{liu2024probing}, the bias in years gets eliminated (they use abstracts, which rarely contain dates). Yet, we recover an AUC of $0.60$ with the most predictive words suggesting a change in topics over time (increased use of \textit{`training'} and \textit{`github'} in recent work). 

In one of the initial MIAs evaluated on data collected post-hoc, Meeus et al.~\cite{meeus2024did} even attempted to control for a potential distribution shift in books by ensuring publication year to be similarly distributed between 1850 and 1910 for members and non-members. Yet, the AUC (Gutenberg (full)) remains near-perfect with the most predictive words relating to template text or formatting. We find that the beginning and end of the books from Project Gutenberg include certain templates (e.g. \textit{`This ebook was transcribed`}), which have changed over time, allowing the classifier to distinguish members from non-members. Similarly for BookMIA~\cite{shi2023detecting}, certain formatting styles have changed between members and non-members (e.g. \textit{`did not`} rather than \textit{`didn't`}). 

The distribution shift can be very subtle. We attempt to mitigate the distribution shift due to the template text by excluding the first and last 10\% of the books (Gutenberg (mid)). Table~\ref{tab:bag_of_words} shows, however, that the AUC of the bag of words classifier does decrease from $0.97$, yet remains as high as $0.89$. The set of predictive words shows that books, while originally published in the same time period but released online at a later stage, contain less old formatting and notably, less politically incorrect words such as the n-word. 

Overall, these results show that a simple bag-of-words classifier can consistently distinguish between members and non-members across all datasets. This implies that a fundamental assumption in membership inference, i.e. the members and non-members are IID, does not hold and that the post-hoc collection of members and non-members fails to provide a proper privacy game to evaluate MIAs. We are hence unable to exclusively attribute the high MIA AUC reported by studies relying on post-hoc data collection~\cite{meeus2024did,shi2023detecting,zhang2024min,xie2024recall,liu2024probing,ye2024data,zhang2024adaptive,wang2024recall,mozaffari2024semantic,kaneko2024sampling} to memorization of the LLM. 

\subsection{Perpetuating flaws in subsequent research}

Since the two initial pieces of work relying on data collected post-hoc~\cite{shi2023detecting,meeus2024did}, we identify at least 10 novel attack methodologies evaluated using datasets collected post-hoc~\cite{sadasivan2024fast,zhang2024min,xie2024recall,liu2024probing,duarte2024cop,ye2024data,zhang2024adaptive,kaneko2024sampling,wang2024recall}, the latest of which proposed (shortly before submission) in September '24~\cite{wang2024recall}. These works either use datasets released by Shi et al.~\cite{shi2023detecting} (WikiMIA and BookMIA) or new datasets of members and non-members created in a similar, post-hoc manner: ArXivMIA~\cite{liu2024probing}, StackMIA~\cite{ye2024data}, Dolma-Book~\cite{zhang2024adaptive} and Wikipedia Cutoff~\cite{mozaffari2024semantic} (the latter two of which have not been made available).

With the results of our bag of words classifier (Table~\ref{tab:bag_of_words}), we find that all (accessible) datasets collected post-hoc suffer from distribution shifts between members and non-members. Therefore, we can not reliably determine whether an MIA achieves a certain performance because it identifies which data points are memorized by the target model or simply due to its ability to detect the distribution shift. Likewise, any comparison of MIA methodologies evaluated with these datasets becomes unreliable.

The problem also extends beyond the development of new MIAs, with subsequent work using datasets collected post-hoc to evaluate the effectiveness of watermarking techniques~\cite{kirchenbauer2023watermark,wei2024proving} as a privacy-preserving technology~\cite{panaitescu2024can} or the success of machine unlearning~\cite{yao2024machine}. The use of flawed MIA evaluation setups in these contexts is highly concerning. Indeed, it could lead to false conclusions about the effectiveness of privacy-preserving technologies, which could have far-reaching implications for LLM development and deployment.


\section{Randomization for MIAs against LLMs}
\label{sec:llm_specific}

Before diving into the potential solutions for proper MIA evaluation, we examine the inherent challenges of creating a randomized setup for MIA evaluation. 

\subsection{Why is it different than for MIAs against ML models?} 

MIAs have been extensively applied to various domains of ML, including image classification~\cite{shokri2017membership,carlini2022privacy,salem2018ml,truex2019demystifying,liu2022membership}, synthetic data generation~\cite{hayes2017logan,stadler2022synthetic,meeus2023achilles} and natural language processing~\cite{shejwalkar2021membership,henderson2018ethical,thomas2020investigating,song2020information}, all evaluated in randomized privacy games. In the \emph{fixed record} privacy game~\cite{ye2022enhanced}, multiple ML models are trained with the target record either included or excluded, commonly referred to as \emph{shadow modeling}~\cite{long2018understanding,long2020pragmatic,stadler2022synthetic,jagielski2020auditing,carlini2022membership,sablayrolles2019white,yeom2018privacy}. For instance, Shokri et al.\cite{shokri2017membership} used up to $100$ shadow models for image classification models, while Song et al.\cite{song2019auditing} trained $10$ shadow models to evaluate privacy in smaller language models (RNNs). However, replicating fixed record privacy games for LLMs is impractical due to the extremely high costs of pretraining.

As a result, privacy games to evaluate MIAs against LLMs likely rely on the \emph{fixed model} scenario, in which a single ML model is trained with randomly sampled target records from the same distribution, either included or excluded. Yet, the practical implementation of fixed model privacy games for LLMs comes with its limitations, which we will address below.

\subsection{Unit of interest to infer membership} 
\label{sec:unit_of_interest}

In traditional MIAs against ML models, the unit of interest to infer membership is naturally defined (e.g. single record, image). For text, most prior work considered sequences of a fixed, often truncated length as the unit of interest. For instance, Carlini et al.~\cite{carlini2021extracting} use sequences of $256$ tokens, while Shi et al.~\cite{shi2023detecting} and subsequent work~\cite{zhang2024min,xie2024recall,liu2024probing,ye2024data,zhang2024adaptive,wang2024recall,mozaffari2024semantic,kaneko2024sampling} consider random excerpts of $32$ to $256$ tokens. When injecting randomized sequences, prior work has considered sequences up to $80$~\cite{wei2024proving} or $100$~\cite{meeuscopyright} tokens. 

In some cases, however, the goal of applying MIAs against LLMs is to infer membership of larger pieces of text, from the level of a document~\cite{meeus2024did,shi2023detecting} to an entire dataset~\cite{maini2024llm}. For instance, in the context of copyright, the goal is to infer whether the entire content (e.g. news articles, book) was used rather than certain excerpts. This has motivated prior work~\cite{meeus2024did,shi2023detecting,maini2024llm} to instantiate document- or even dataset-level membership inference, aiming to aggregate membership signal to larger pieces of text. 

Formally, a document-level MIA aims to determine whether a given document $d_t = (t_1, t_2, ..., t_{N_{d_i}})$ containing $N_{d_t}$ tokens, was included in the training dataset $\mathcal{D}$ for target LLM $\mathcal{M}$. The MIA defines a scoring function $f_{\text{doc}}(\mathcal{M}, d_t)$ such that $\hat{y}_{\text{doc}}(\mathcal{M}, d_t) = 1 \text{ if } f_{\text{doc}}(\mathcal{M}, d_t) \geq \tau_{\text{doc}} \text{, and } 0 \text{ otherwise}$. The true membership label is $y_{\text{doc}}(\mathcal{M}, d_t) = 1 \text{ if } d_i \in \mathcal{D} \text{, and } 0 \text{ otherwise}$. For simplicity, we argue that the task of dataset inference as proposed by Maini et al.~\cite{maini2024llm} is a generalization of document-level MIAs.

Document-level MIAs differ from sequence-level MIAs against LLMs, as aggregating meaningful signal across significantly more tokens might be challenging. Multiple ways have been proposed to compute $f_{\text{doc}}(\mathcal{M}, d_t)$. Meeus et al.~\cite{meeus2024did} iterates through the entire document $d_t$, queries the LLM on truncated chunks, and aggregates all token-level probabilities to multiple document-level features as input for a membership classifier. Shi et al.~\cite{shi2023detecting} proposes to sample $k$ random excerpts, i.e. sequences  $s_{tj}$, from document $d_t$ to compute the sequence-level scoring function $f_{\text{seq}}(\mathcal{M}, s_{tj})$ for all $j=1, \ldots k$. This document-level scoring function is then computed as an average binary prediction $y_{\text{doc}}(\mathcal{M}, d_t) = \frac{1}{k} \sum_{j=1}^k \hat{y}_{\text{seq}}(\mathcal{M}, s_{tj})$, where $\tau_{\text{seq}}$ is determined to maximize accuracy on the sequence-level MIA. Lastly, Maini et al.~\cite{maini2024llm} aggregates values of the sequence-level scoring function to a document-level statistical test.

\subsection{Overlap between members and non-members} 
\label{sec:overlap}

LLMs require massive datasets for pre-training, often collected from diverse internet sources (e.g. 15 trillion tokens for LLaMA-3~\cite{llama3modelcard}). Since text is composed of sequences of tokens from a finite vocabulary, it tends to be lower-dimensional and more fragmented compared to data types like images. Given the sheer scale of these datasets, there is likely significant duplication of certain sequences. Moreover, text from the internet likely naturally contains significant duplication, with excerpts from e.g. books or news articles frequently appearing across various sources. 

This inherent overlap of text is a well-known issue which has been heavily studied in the context of properly evaluating newly released LLMs. For instance, in GPT-3~\cite{brown2020language}, authors found a significant overlap between the training data and established evaluation benchmarks. To evaluate how the model generalizes to unseen data, they aggressively deduplicate ($13$-gram overlaps) between the training data and benchmarks. This methodology (\emph{decontamination}) has since been adopted and further refined by subsequent model developers~\cite{wei2021finetuned,du2022glam,touvron2023llama2}. However, even these exact deduplication methods remain insufficient to address near-duplicates, semantic duplicates or translations~\cite{yao2024datacontaminationcrosslanguage}, highlighting the persistent challenge of effective decontamination. 

The overlap also poses similar challenges for MIAs. Indeed, non-member data that is partially seen during training may act as quasi-members, potentially making the membership inference task poorly defined. Moreover, recent research suggests that the inclusion of near-duplicates of members in the training data enhances the members' distinguishability in MIAs~\cite{shilov2024mosaic}, further complicating the inference process.

Recent works~\cite{li2023mope,duan2024membership,maini2024llm} have proposed to use the randomized split of train and test data from the Pile~\cite{pile} to evaluate MIAs against LLMs. Indeed, these splits have been used to train and validate some open-source models~\cite{biderman2023pythia,gpt-neo} and thus enable to instantiate a proper, fixed-model privacy game. However, Duan et al.~\cite{duan2024membership} finds that these splits, just as encountered for evaluation benchmarks, can indeed have substantial overlap, making it harder for MIAs to succeed. They report an average of 32.5\% $7$-gram overlap between members and non-members for Wikipedia articles from the Pile~\cite{pile}, even reaching up to 76.9\% overlap for data collected from Github.

\section{Towards proper MIA evaluation}
\label{sec:solution_mias}

Recognizing the importance and challenges of creating a randomized setup for evaluating MIAs against LLMs, we review and discuss potential solutions from the literature.

\subsection{Randomized test split}
\label{sec:pythia_solution}

The privacy game for a fixed model~\cite{ye2022enhanced} only requires one training run, and could be naturally instantiated alongside regular training. When planned and established before training begins, this remains feasible even for expensive LLMs. 

Some pretrained LLMs make their entire training dataset open-source, including a randomized and controlled split of `test' data. While this set of test documents is primarily intended for a fair evaluation of the model's ability to generalize its language modeling capabilities to unseen language, such test sets might prove very valuable to study and evaluate MIAs~\cite{duan2024membership,li2023mope,maini2024llm}. The training and validation splits have been, for instance, made available for models such as Pythia~\cite{biderman2023pythia}, GPT-NEO~\cite{gpt-neo}, OLMo~\cite{OLMo} and CroissantLLM~\cite{faysse2024croissantllm}. 

First, we use our bag of words classifier to analyze whether the distribution shift identified earlier for members and non-members collected post-hoc, is now eliminated in a real-world, randomized set of members and non-members. We consider three such datasets. First, we consider $1,000$ members and non-members directly sampled from the train and test splits of (the uncopyrighted version of) The Pile~\cite{pile_uncopyrighted} (used to train and validate Pythia~\cite{biderman2023pythia} and GPT-NEO~\cite{gpt-neo}). Next, we consider both versions of the MIMIR dataset (13\_0.8 and 7\_0.2)~\cite{duan2024membership}. As discussed in Sec.~\ref{sec:overlap}, Duan et al.~\cite{duan2024membership} identified a substantial overlap between members and non-members in the randomized split from The Pile, making the notion of membership poorly defined and making it harder for MIAs to succeed. To mitigate this issue, they apply a deduplication strategy on the non-member data, removing non-members with more than 80\% overlap in $13$-grams (13\_0.8) and more aggressively, samples with more than 20\% overlap in $7$-grams (7\_0.2). Authors release the result as the MIMIR dataset. 

\begin{table}[t]
    \centering
    \caption{AUC (mean$\pm$standard deviation) for the bag of words classifier applied to subsets of The Pile, either directly sampled from its uncopryighted version~\cite{pile_uncopyrighted} or from  MIMIR~\cite{duan2024membership} (which applied further deduplication).}
    \begin{tabular}{ccccc}
    \toprule
         & & MIMIR~\cite{duan2024membership} & MIMIR~\cite{duan2024membership}\\
        Subset & The Pile~\cite{pile_uncopyrighted} & (13\_0.8) & (7\_0.2)\\
        \midrule 
        \midrule 
         Wikipedia & $0.497 \pm .017$ & $0.516 \pm .040$ &\bm{$0.580\pm.030$}\\
         \midrule 
         Pile CC  & $0.551 \pm .026$ & $0.525 \pm .028$ & $0.539\pm.038$\\ 
         \midrule 
         PubMed & $0.481 \pm .010$ & $0.508 \pm .024$ & \bm{$0.783\pm.017$}\\ 
         \midrule 
         ArXiv & $0.502 \pm .017$ & $0.501 \pm .039$ & \bm{$0.730\pm.013$}\\ 
         \midrule 
         DM Math & $0.503 \pm .028$ & $0.499 \pm .025$ & \bm{$0.799\pm.063$} \\
         \midrule 
         HackerNews & $0.488 \pm .020$ & $0.497 \pm .021$ & \ding{55} \\ 
         \midrule 
         Github & $0.507 \pm .014$ & \bm{$0.679 \pm .028$} & \bm{$0.860\pm.035$} \\ 
         \midrule 
        \bottomrule \\
    \end{tabular}
    \label{tab:bag_of_words_pile}
\end{table}

The bag of words performance in Table~\ref{tab:bag_of_words_pile} (first column) first shows that, for the data directly sampled from the train and test split across all $7$ subsets from the Pile, the bag of words classifier barely performs better than a random guess; consistent with what we expect from a randomized split.

For the MIMIR dataset~\cite{duan2024membership}, however, the results (second and third columns) are more nuanced. For the moderate level of deduplication (13\_0.8), we find the AUC to remain close to a random guess baseline for almost all data subsets. This means that the deduplication did not introduce a distribution shift detected by the bag of words classifier, while removing non-members with substantial overlap in the training dataset. For data collected from Github, however, our bag of words classifier obtains an AUC of $0.68$, suggesting that there is a meaningful distribution shift between members and non-members. This implies that the overlap between members and non-members for data collected from Github is so substantial that the non-members remaining after deduplication are mostly outliers, which initiates the distribution shift. Duan et al.~\cite{duan2024membership} indeed confirms that the overlap is significantly more substantial for Github than for other data subsets and illustrates specific outliers that remain after deduplication. For the more aggressive deduplication (7\_0.2), we find similarly large AUC for almost all subsets. \textbf{This suggests that deduplicating too aggressively post-hoc might distort the exact properties of a randomized setup that are desired to evaluate MIAs.} Future research is needed to understand the level of deduplication which can be applied post-hoc to mitigate the effect of the overlap issue, while not introducing any distribution shift. Our results suggest that MIMIR (13\_0.8), apart from the data from Github, would serve as a randomized setup to evaluate MIAs against LLMs. However, we note again that the AUC achieved by the bag of words classifier only represents a lower bound; if its performance is close to a random guess, this does not necessarily imply the absence of any distribution shift. 

We instantiate a range of MIAs from the literature on the MIMIR dataset~\cite{duan2024membership} (13\_0.8, and not considering the Github subset) using Pythia-dedup-6.9B~\cite{biderman2023pythia}. Table~\ref{tab:pile_auc} shows that the original attacks~\cite{yeom2018privacy,carlini2021extracting,mattern2023membership,shi2023detecting} do not perform significantly better than random, and neither do any of the $5$ recently proposed attacks~\cite{zhang2024min,wang2024recall,ye2024data,zhang2024adaptive,kaneko2024sampling} which allegedly outperformed the original ones. 

We also instantiate document-level MIAs on a randomized split of entire ArXiv documents from the Pile~\cite{pile_uncopyrighted}. We consider both MIAs applied to all sequences in the document, scaled to the document-level as proposed by Shi et al.~\cite{shi2023detecting} (formalized in Sec.~\ref{sec:unit_of_interest}) as well as the document-level MIA (using a meta-classifier) proposed by Meeus et al.~\cite{meeus2024did}. Also for document-level MIAs, we find the MIA AUC to barely perform any better than random. We provide the results in Table~\ref{tab:doc_level} with the detailed experimental setup in Appendix B.

\begin{table*}[ht]
    \centering
    \caption{Benchmarking MIAs on the MIMIR dataset~\cite{duan2024membership} and Pythia-dedup-6.9B~\cite{biderman2023pythia}. Sequences are sampled from the train (members) and test (non-members) split of The Pile~\cite{pile}. MIA AUC computed using 500 members and 500 non-members.}
        \begin{tabular}{lcccccc}
    \toprule
          & \multicolumn{6}{c}{Subset of The Pile~\cite{pile}} \\
         MIA & Wikipedia & Pile CC & PubMed Central & ArXiv & DM Math & HackerNews\\
         \midrule
         \midrule
         Bag of words (model-less) & $0.516 \pm .040$ & $0.525 \pm .028$ & $0.481 \pm .010$ & $0.502 \pm .017$ & $0.503 \pm .028$ & $0.488 \pm .020$\\
         \midrule
         \midrule
         \textit{Loss}~\cite{yeom2018privacy} & $0.515 \pm .013$ & $0.512 \pm .012$ & $0.504 \pm .013$ & $0.525 \pm .013$ & $0.484 \pm .013$ & $0.517 \pm .013$\\ 
         \cmidrule{1-7}
        \textit{Zlib}~\cite{carlini2021extracting} & $0.516 \pm .013$ & $0.512 \pm .013$ & $0.506\pm.013$ & $0.524\pm.013$ & $0.486\pm.013$ & $0.517\pm.013$ \\ 
         \cmidrule{1-7} 
        \textit{Lower}~\cite{carlini2021extracting}& $0.518\pm.013$ & $0.519\pm.013$ & $0.513\pm.012$ & $0.518\pm.012$ & $0.493\pm.013$ & $0.507\pm.012$ \\ 
         \cmidrule{1-7}
        \textit{Ratio-LLaMA-2}~\cite{carlini2021extracting}& $0.529\pm.013$ & $0.516\pm.013$ & $0.524\pm.013$ & $0.537\pm.013$ & $0.503\pm.013$ & $0.535\pm.013$ \\ 
         \cmidrule{1-7}        
         \textit{Neighborhood}~\cite{mattern2023membership}& $0.501\pm.013$ & $0.511\pm.014$ & $0.496\pm.013$ & $0.530\pm.013$ & $0.511\pm.013$ & $0.484\pm.013$ \\ 
        \cmidrule{1-7}
        \textit{Min-K\% Prob}~\cite{shi2023detecting} & $0.515\pm.013$ & $0.519\pm.012$ & $0.510\pm.013$ & $0.527\pm.013$ & $0.492\pm.013$ & $0.530\pm.013$ \\
        \cmidrule{1-7}
        \textit{Min-K\%++}~\cite{zhang2024min} & $0.510\pm.013$ & $0.516\pm.013$ & $0.506\pm.013$ & $0.520\pm.012$ & $0.491\pm.013$ & $0.529\pm.013$ \\
        \cmidrule{1-7}
        \textit{PAC}~\cite{ye2024data} & $0.482\pm.013$ & $0.476\pm.013$ & $0.479\pm.013$ & $0.477\pm.013$ & $0.505\pm.013$ & $0.485\pm.013$ \\
        \cmidrule{1-7}
        \textit{SURP}~\cite{zhang2024adaptive} & $0.511\pm.013$ & $0.507\pm.013$ & $0.504\pm.013$ & $0.508\pm.013$ & $0.492\pm.013$ & $0.522\pm.013$ \\
        \cmidrule{1-7}
        \textit{SaMIA}~\cite{kaneko2024sampling} & $0.468\pm .014$ & $0.521\pm.012$ & $0.518\pm.013$ & $0.495\pm .013$ & $0.493\pm .013$ & $0.539\pm .013$ \\
        \cmidrule{1-7}
        \textit{SaMIA*zlib}~\cite{kaneko2024sampling} & $0.507\pm .012$ & $0.514\pm.012$ & $0.527\pm.013$ & $0.491\pm.013$ & $0.524\pm .013$ & $0.539\pm.013$ \\
         \midrule
         \bottomrule \\
    \end{tabular}
\label{tab:pile_auc}
\end{table*}

\textbf{Limitations.} While randomized, held-out test splits (when available) generally provide a proper fixed-model privacy game, the substantial overlap between members and non-members remains a core challenge. Post-hoc deduplication could potentially mitigate the overlap, yet determining the appropriate level of deduplication is difficult to evaluate and might distort the randomization. Its effectiveness likely varies depending on the dataset type, as demonstrated in the case of the GitHub subset. Utilizing the bag of words classifier to guide the appropriate level of deduplication could help achieve the right trade-off in practice. 

The practical significance of the overlap issue is a nuanced, perhaps philosophical, question. To rigorously examine what an LLM memorizes and how effectively MIAs can distinguish between members and non-members, one might prefer a setup with minimal overlap. Yet, overlap remains an inherent feature of text and it might be appropriate to account for overlap when inferring membership in real-world applications. For instance, when assessing whether a copyright-protected book was used to train an LLM, well-known quotes from it may be widespread online and included as near-duplicates in the training data. To identify whether the book was used in its entirety to train the model, it is likely essential to also account for such overlap. Whether MIAs, to e.g. study privacy risk, or detect copyright violations in practice, should be evaluated on datasets with an overlap thus remains an open question. 

\subsection{Injection of randomized unique sequences}

To study sequence-level memorization, instead of relying on a randomized test split of documents, prior work has also proposed the injection of highly unique sequences. 

For smaller language models, the controlled injection of specific sequences in the training data has been studied extensively. Henderson et al.~\cite{henderson2018ethical} injects a variety of key-value pairs in a sequence-to-sequence dialogue model, while Carlini et al.~\cite{carlini2019secret} proposes the injection of \emph{canaries}, unique sequences of random numbers, in the training set of LSTM models; both to measure how these sequences are memorized. Further work has, for instance, injected canaries to study memorization of federated learning~\cite{thakkar2020understanding} or of word embeddings~\cite{thomas2020investigating}. 

More recently, Wei et al.~\cite{wei2024proving} investigates how data watermarks, i.e. random sequences of tokens injected into the training dataset, are memorized by small scale language models (70M parameters trained on 100M tokens). Further, Meeus et al.~\cite{meeuscopyright} injected canaries (called \emph{traps sequences}) into the pretraining dataset of CroissantLLM~\cite{faysse2024croissantllm}, a 1.3B parameter model trained on 3T tokens. The canaries are synthetically generated, unique sequences, injected with varying sequence length, number of repetitions, and perplexity. As the canaries are synthetically generated, by sampling from a pretrained LLM, the same process allows generating new, unique canaries coming from the exact same distribution, yet not part of the training dataset of CroissantLLM~\cite{faysse2024croissantllm} (non-members). Lastly, Zhang et al.~\cite{zhang2024membership} also states that inferring membership of specially crafted canaries offers a viable path for proving whether LLMs are trained on certain data. 

While Meeus et al.~\cite{meeuscopyright} use this setup to study how the inclusion of canaries enables content owners to track whether their content has been used for LLM training, the setup also provides a randomized and controlled setup to evaluate sequence-level MIAs against LLMs. We confirm this to be the case by fitting the bag-of-words classifier on copyright traps (using $500$ members and non-members), recovering an AUC of $0.543 \pm 0.037$ for sequences of $100$ tokens. 

Compared to the setup available with a randomized test split (Sec.~\ref{sec:pythia_solution}), the use of randomized unique sequences to evaluate MIAs might have distinct benefits. 

First, this setup is designed for sequences and eliminates the need to sample sequences from train and test splits. 

Further, when the randomized sequences are highly unique, they could effectively eliminate the challenge of overlap between members and non-members. Meeus et al.~\cite{meeuscopyright} generate synthetic sequences of high perplexity, suggesting that these sequences would likely have less overlap with the rest of the target model training data than a randomized test split.  

Lastly, injected sequences can be repeated. As it is well established that repeated sequences are memorized more~\cite{carlini2022quantifying,kandpal2022deduplicating}, injecting sequences with a varying number of repetitions would provide an MIA evaluation setup across multiple regimes of LLM memorization. This could enable the comparison of different MIAs with (more) statistical significance.

We instantiate MIAs from the literature on the dataset of member and non-member canaries as made available by Meeus et al.~\cite{meeuscopyright} on the open-source CroissantLLM~\cite{faysse2024croissantllm}. Table~\ref{tab:traps_auc} summarizes the MIA AUC across a number of repetitions. For $1,000$ repetitions, we recover MIAs reaching a performance significantly better than a random guess, with \textit{Ratio-LLaMA-2}~\cite{carlini2021extracting} (AUC of $0.70$) and \textit{Min-K\% Prob}~\cite{shi2023detecting} (AUC of $0.63$) being the best performing methods. 

\begin{table}[ht]
    \centering
    \caption{Benchmarking MIAs across copyright traps for CroissantLLM~\cite{faysse2024croissantllm}: MIA AUC for $500$ members and non-members.}
    \begin{tabular}{lccc}
    \toprule
        & \multicolumn{3}{c}{$n_\text{rep}$} \\
        MIA & 10 & 100 & 1000 \\
        \midrule
        \midrule
        Bag of words
        & $0.543 \pm 0.37$ & $0.543 \pm 0.37$ & $0.543 \pm 0.37$  \\ 
        \midrule
        \midrule
        \textit{Loss}~\cite{yeom2018privacy} 
        & $0.509\pm.019$ & $0.538\pm.018$ & $0.592\pm.019$  \\ 
        \cmidrule{1-4}
         
        \textit{Zlib}~\cite{carlini2021extracting} 
        & $0.514\pm.019$ & $0.537\pm.019$ & $0.569\pm.017$ \\ 
        \cmidrule{1-4} 
         
        \textit{Lower}~\cite{carlini2021extracting}
        & $0.526\pm.018$ & $0.548\pm.018$ & $0.613\pm.018$ \\ 
        \cmidrule{1-4}
         
        \textit{Ratio-LLaMA-2}~\cite{carlini2021extracting}
        & $0.542\pm.018$ & $0.617\pm.018$ & $0.704\pm.017$ \\ 
        \cmidrule{1-4}
         
        \textit{Neighborhood}~\cite{mattern2023membership}
        & $0.488\pm.019$ & $0.512\pm.019$ & $0.566\pm.018$\\ 
        \cmidrule{1-4}
        
        \textit{Min-K\% Prob}~\cite{shi2023detecting} 
        & $0.509\pm.018$ & $0.547\pm.018$ & $0.626\pm.018$\\
        \cmidrule{1-4}
        
        \textit{Min-K\%++}~\cite{zhang2024min} 
        & $0.514\pm.018$ & $0.531\pm.018$ & $0.599\pm.018$\\
        \cmidrule{1-4}
        
        \textit{ReCaLL}~\cite{xie2024recall} 
        & $0.508\pm.019$ & $0.538\pm.018$ & $0.591\pm.018$\\
        \cmidrule{1-4}
        
        \textit{Probe (real)}~\cite{liu2024probing} 
        & $0.482\pm.018$ & $0.460\pm.019$ & $0.480\pm.018$\\
        \cmidrule{1-4}
        
        \textit{PAC}~\cite{ye2024data} 
        & $0.512\pm.017$ & $0.545\pm.018$ & $0.587\pm.017$\\
        \cmidrule{1-4}
        
        \textit{SURP}~\cite{zhang2024adaptive} 
        & $0.490\pm.018$ & $0.550\pm.018$ & $0.592\pm.019$\\
        \cmidrule{1-4}
        
        \textit{CON-ReCall}~\cite{wang2024recall} 
        & $0.504\pm.018$ & $0.529\pm.018$ & $0.477\pm.019$\\
        \cmidrule{1-4}
        
        \textit{SaMIA}~\cite{kaneko2024sampling} 
        & $0.524\pm.018$ & $0.582\pm.018$ & $0.578\pm.018$\\
        \cmidrule{1-4}
        
        \textit{SaMIA*zlib}~\cite{kaneko2024sampling} 
        & $0.578\pm.018$ & $0.464\pm.019$ & $0.538\pm.019$\\
        \midrule
        \bottomrule \\
    \end{tabular}
\label{tab:traps_auc}
\end{table}

\textbf{Limitations.} First, the MIA performance might only represent a worst-case risk and be not representative of real-world data. Indeed, highly unique sequences mitigate any overlap between members and non-members, making it easier for MIAs to succeed, especially when repeated during training. Moreover, Meeus et al.~\cite{meeuscopyright} shows that sequences with larger perplexity are more vulnerable, and considers sequences with a perplexity higher than naturally occurring ones. Hence, evaluating MIAs on unique sequences might give a worst-case MIA risk compared to real-world data. While this might be beneficial for some use-cases (e.g. privacy auditing of methods with formal guarantees~\cite{long2018understanding,long2020pragmatic,stadler2022synthetic}, detecting copyright-protected content~\cite{meeuscopyright,wei2024proving}), evaluating MIAs on (artificially) unique sequences might not be representative of risks in practice. Notably, MIAs are likely to advance over time, and focusing on higher-risk sequences could establish a safety margin. 

Second, the success of MIAs might depend on certain properties of the sequence. For instance, the \textit{Ratio} MIA~\cite{carlini2021extracting} relies on the use of a reference model, which should normalize the membership score by how `complex' a certain sequence is compared to other natural language. Surely, the extent to which this normalization is effective for an MIA against a target LLM depends on the choice of reference model, its training data and how well it generalizes to unseen data. Meeus et al.~\cite{meeuscopyright} use LLaMA-2 7B~\cite{touvron2023llama2} to synthetically generate the trap sequences, likely affecting the success of certain reference models in detecting such traps. Also Duan et al.~\cite{carlini2021extracting} finds \emph{`choosing
the right reference model for a target LLM challenging and largely empirical'}. Moreover, we hypothesize that certain MIAs may vary in effectiveness based on sequence length (e.g. Min-K\% Prob~\cite{shi2023detecting}, which relies on a subset of token-level probabilities). Similarly, some MIAs (e.g. \textit{SMIA}~\cite{mozaffari2024semantic}) depend on the semantic meaning of a target sequence, which might be less effective for sequences with very high perplexity due to the lack of semantically meaningful embeddings in such cases. Using randomized unique sequences as an evaluation setup could thus influence the relative performance of MIAs.

\subsection{Finetuning}

Thus far, we have worked under the assumption that training one (or more) models for the purpose of MIA evaluation is beyond reach for LLMs. While this is the case at the scale of modern LLM pretraining, it is generally feasible to run continued pretraining or to finetune LLMs on smaller datasets, and potentially using parameter-efficient finetuning methods~\cite{hu2021lora, hyeonwoo2023fedparalowrankhadamardproduct, zhang2023adaloraadaptivebudgetallocation, dettmers2024qlora, buehler2024xloramixturelowrankadapter, liu2024doraweightdecomposedlowrankadaptation}. Researchers could thus finetune LLMs, with controlled injection of randomized members and non-members, to then evaluate MIAs on the finetuned models (fixed model) or even finetune multiple LLMs for a certain record (fixed record).

For instance, Mattern et al.~\cite{mattern2023membership} finetunes GPT-2~\cite{radford2019language} to evaluate their MIA on a controlled set of members and non-members. Mireshghallah et al.~\cite{mireshghallah2022empirical} also finetunes GPT-2 models~\cite{radford2019language} and finds that finetuning the head of the model leads to better attack performance. Other work finetuned models to measure data extraction~\cite{zeng2023exploring}, or to apply MIAs for plagiarism detection~\cite{lee2023language} and studying PII leakage~\cite{lukas2023analyzing}.

While most of the prior work discussed in this paper consider pretrained LLMs as the target model, we believe that leveraging finetuning for MIA evaluation merits a discussion as part of the potential solutions. In particular, we find that new MIA methods proposed in recent work are reported to outperform prior methods~\cite{zhang2024min,xie2024recall,ye2024data,zhang2024adaptive,wang2024recall,mozaffari2024semantic}, while they are all evaluated on flawed setups using pretrained LLMs. Leveraging finetuning for MIA evaluation could enable a computationally feasible way for more reliable benchmarking. We further identify two additional benefits. 

First, similar to the injection of unique sequences with varying repetitions, finetuning could be done for varying memorization regimes (adjusting e.g. number of parameters updated during finetuning, dataset size, learning rate~\cite{mireshghallah2022empirical}). As such, this setup would be well suited to compare MIA performance across methods.  

Further, from a practical perspective, it is very sensible to study MIAs in posttraining attack models. Finetuning (or continued pretraining) is often done on smaller, domain-specific, high quality datasets which are often more susceptible to contain sensitive information. It is cost effective to start off from generalist model checkpoints (e.g. LLaMA-3~\cite{llama3modelcard}, Mistral-7B~\cite{jiang2023mistral}) and train on a domain-specific, causal language modeling task for a few billions of tokens to specialize models. Examples include finetuning for legal domain adaptation~\cite{colombo2024saullm7bpioneeringlargelanguage}, for multilingual abilities~\cite{alves2024toweropenmultilinguallarge}, and for medical domain adapation~\cite{labrak2024biomistralcollectionopensourcepretrained}. Moreover, annealing, or post-training techniques in which LLMs are trained on high-quality data splits after their initial pretraining phase, is now understood to be essential to boost model performance~\cite{hu2024minicpmunveilingpotentialsmall, llama3modelcard}. Such high-quality data (news articles, textbooks, synthetic data) is often the category most likely to be copyright-protected.

\textbf{Limitations.} For some use-cases, including detecting the use of copyright-protected content or evaluation benchmarks during pretraining, the evaluation of MIAs remains desired for pretrained rather than for finetuned models. How MIA evaluation on finetuned models translates to larger models, datasets or other hyperparameters remains an open question~\cite{biderman2024emergent}. 

\subsection{Post-hoc control methods}

Throughout this work, we have made the case for randomized setups to evaluate MIAs against LLMs. We identified as potential solutions: (i) randomized test split, (ii) injection of randomized unique sequences and (iii) finetuning.  However, such randomized setups are generally unavailable for most LLMs, especially for the widely used recent LLMs, such as Mistral-7B~\cite{jiang2023mistral}, LLaMA-2/3~\cite{touvron2023llama2,llama3modelcard} and GPT-4~\cite{gpt4techreport}.

Therefore, we examine the options to still apply MIAs on real-world LLMs for which a controlled setup is unavailable. This might be useful to compare the memorization of certain LLMs or to calibrate MIAs (e.g. determining the threshold $\tau_{\text{seq}}$) before making an actual inference.

\textbf{Beating the model-less baseline.} A potential solution is to revisit the baseline used to infer binary membership. MIAs against ML models traditionally use a 50/50 split of membership for evaluation, resulting in a random guess AUC of $0.5$~\cite{shokri2017membership,carlini2022membership,choquette2021label,li2022auditing,salem2018ml,truex2019demystifying,hayes2017logan,leino2020stolen,nasr2018machine,sablayrolles2019white,yeom2018privacy}. Alternatively, Jayaraman et al.~\cite{jayaraman2020revisiting} argues that a balanced prior for membership is not realistic in practice and proposes to evaluate MIAs with skewed priors. More recently, Kazmi et al.~\cite{kazmi2024panoramia} proposes privacy auditing of ML models using a baseline from a (model-less) classifier. Further, Chang et al.~\cite{chang2024context} evaluate a bag of words classifier on the MIMIR dataset~\cite{duan2024membership} and use this as a baseline for their new MIA. 

Hence, we could set the performance of the bag-of-words classifier as a prior. An effective MIA methodology should then outperform this baseline by leveraging access to the target model exploiting its memorization of members. 

This solution has significant limitations in practice. First, the performances of the bag of words classifier applied to the widely used datasets collected post-hoc are near-perfect (Table~\ref{tab:bag_of_words}). With such a significant distribution shift between members and non-members, it becomes practically impossible for novel MIAs to surpass this baseline and might lead us to (incorrectly) conclude that the target model does not memorize. Second, such model-less baselines will always be a lower bound. In Sec.~\ref{sec:flawed}, we only used a simple bag-of-words classifier, since its performance was already substantial. Had this classifier performed closer to random guessing, a subtler distribution shift might still exist, potentially detectable by more sophisticated classifiers~\cite{zhang2015character,devlin2018bert}. 

\begin{table*}[t]
    \centering
    \caption{Recommended benchmarks to evaluate new MIAs against LLMs proposed in the future, highlighting benefits and limitations.}
    \begin{tabular}{C{1.2cm}|p{6cm}|C{2cm}|p{3.3cm}|p{3.3cm}}
    \toprule
        Unit of &  &  &  &  \\
        interest & Dataset & Model(s) & Benefits & Limitations \\
        \midrule
        \midrule
        Sequence & The randomized split of The Pile~\cite{pile}, with additional deduplication, as made available by Duan et al.~\cite{duan2024membership} in the MIMIR dataset\footnote{\url{https://huggingface.co/datasets/iamgroot42/mimir}}. We recommend using the split where sequences with more than 80\% of $13$-gram overlapping are not removed, not considering data from Github. & Suite of Pythia~\cite{biderman2023pythia} and GPT-NEO~\cite{gpt-neo} models. & Allows for randomized MIA evaluation. & MIAs are more challenging due to the inherent overlap between members and non-members This however represents a more realistic setup in general. \\
        \midrule 
        Sequence & The randomized unique \emph{traps sequences} as made available by Meeus et al.~\cite{meeuscopyright}, across all number of repetitions $(10, 100, 1000)$. & CroissantLLM~\cite{faysse2024croissantllm} & Allows for randomized MIA evaluation in a real-world model, limiting the overlap between members and non-members, and across different memorization regimes. & MIA performance might not be realistic, and might be specific to sequence characteristics such as length, and perplexity. \\
         \midrule
         \midrule
        Document & Randomized set of ArXiv papers sampled from the train and test splits from The Pile~\cite{pile}, both available as complete documents as well as in sequences. & Suite of Pythia~\cite{biderman2023pythia} and GPT-NEO~\cite{gpt-neo} models. & Allows for randomized MIA evaluation. & MIAs are more challenging due to the inherent overlap between members and non-members This however represent a more realistic setup in general. \\
        \midrule
        Document & Set of ArXiv papers sampled close to the training data cut-off date for OpenLLaMA\cite{openlm2023openllama}. & Suite of OpenLLaMA\cite{openlm2023openllama} models. & Allows for MIA evaluation on data collected post-hoc with any distribution shift between members and non-members largely mitigated (RDD). & Might still be subject to a distribution shift, and faces potential data overlap. \\
        \midrule 
    \bottomrule 
    \end{tabular}
\label{tab:benchmark_proposal}
\end{table*}

\textbf{Regression Discontinuity Design (RDD).} In the domain of causal inference, Regression Discontinuity Design (RDD) is a technique used to study the effect of a treatment when the assignment to the treatment has not been randomized but is instead based on a cutoff point. 
By sampling observations in a principled manner close to this cutoff, RDD enables the estimation of the treatment effect even in the absence of a randomized setup. RDD is for instance widely used in econometrics~\cite{lee2010regression}, where cutoff points might include income levels for welfare programs or geographical borders when studying different policy implications.

We here examine if the RDD technique can be applied to the task of membership inference for LLMs. Specifically, we will sample data samples published close to the cutoff date for the LLM training data, attempting to mitigate any distribution shift between members and non-members. 

We instantiate this RDD setup on OpenLLaMA~\cite{openlm2023openllama}, a model for which the entire training dataset is known. As members, we consider papers from a month before the training data cutoff date (February '23). For non-members, we consider papers released shortly after (March '23). 

First, we apply the same bag of words model to quantify any distribution shift still present in the RDD data. We find an AUC of $0.534 \pm 0.027$ when distinguishing members from non-members, suggesting that by sampling documents close to the cutoff date, we eliminate a substantial part of the previously present distribution shift for papers as used by Meeus et al.~\cite{meeus2024did} (AUC of $0.69$) or for paper abstracts as used in ArXivMIA~\cite{liu2024probing} (AUC of $0.60$).

Table~\ref{tab:doc_level} (Appendix B) summarizes the AUC for both sequence- as well as document-level MIAs across the literature applied to this setup. We find that the performance of MIAs remains close to a random guess baseline, as also found in the case of a randomized test split (Sec.~\ref{sec:pythia_solution}). 

While the RDD method we propose seems to offer a viable solution for the setup at hand (using OpenLLaMa~\cite{openlm2023openllama} and ArXiv papers), its practical implementation may face limitations. First, it assumes knowledge of the training data cut-off date (which is likely quite different from the model release date) and relies on the availability of sufficient data released close to this date. For ArXiv papers, more than $10,000$ new papers are added every month\footnote{\url{https://arxiv.org/stats/monthly_submissions}}, allowing for ample data collection close to any cut-off date. While similar arguments can be made for other frequently updated data sources (e.g. news articles), the RDD approach could prove challenging for less frequently updated sources (e.g. crime novels). Second, the RDD method aims to significantly mitigate distribution shifts between members and non-members, but its quantification faces similar issues as discussed above. Lastly, the RDD setup does not address the issue of potential overlap between members and non-members. This might be preferred if one wants to calibrate realistic MIA performance in practice, but maybe less so when benchmarking new methods. 

\textbf{Debiasing techniques.} Eichler et al.\cite{eichler2024nob} introduced two novel methods to mitigate bias between members and non-members collected post-hoc. The first method creates `non-biased' datasets, where members and non-members share similar n-gram distributions. The second one creates `non-classifiable' datasets, which resist model-less baselines like those used here and by Das et al.\cite{das2024blind}. While authors show these techniques to eliminate biases present in the original datasets collected post-hoc, this approach faces similar limitations as other post-hoc control methods. Fully eliminating known biases, not introducing new biases and accurately evaluating their presence, remains a practical challenge. 

\section{Comparison to concurrent work}
\label{sec:compare_concurrent}

Concurrently with this work (the preprints were released within the same week), Das et al.~\cite{das2024blind} propose to use bag of words classifiers, which they call a `blind' baseline, to evaluate MIAs against LLMs. They show bag of word classifiers to outperform MIAs in two datasets: Shi et al.~\cite{shi2023detecting} and Meeus et al.~\cite{meeus2024did}, discovering significant distribution shifts. As a potential solution, they propose the use of a randomized test split as made available for some LLMs. 

We independently introduce a bag of words classifier, showed them to outperform MIAs, and identified distribution shifts in $6$ datasets~\cite{meeus2024did,shi2023detecting,li2022auditing,ye2024data}, including the three used by Das et al.~\cite{das2024blind}. We further analyzed how post-hoc data deduplication such as employed to create the MIMIR dataset~\cite{duan2024membership} might introduce new biases, and identify a range of recent work on MIAs against LLMs evaluated in flawed setups (Table~\ref{tab:overview}). Lastly, we carefully examine different potential solutions going forward (Sec.~\ref{sec:solution_mias}), evaluate a wide range of MIAs in these setups (Tables~\ref{tab:pile_auc} and~\ref{tab:traps_auc}) and propose a recommended set of benchmarks (Table~\ref{tab:benchmark_proposal}).

\section{Conclusion}

We identified that recent research on MIAs against LLMs often uses flawed evaluation setups. In the last year, research has proposed at least $10$ different MIAs, whose performance has been impossible to fairly compare and benchmark.

After carefully examining potential solutions, we identify a set of benchmarks to use when new MIAs are proposed and summarize our findings in Table~\ref{tab:benchmark_proposal}. For sequence-level MIAs, we propose a combination of the MIMIR dataset~\cite{duan2024membership} (with limited deduplication) and the randomized trap sequences provided by Meeus et al.~\cite{meeuscopyright}. MIMIR enables a randomized setup for MIA evaluation across different model sizes~\cite{biderman2023pythia,gpt-neo} and, despite its overlap issues, serves as a realistic benchmark for MIAs in practice. The trap sequences allow for benchmarking new MIA methodologies across different levels of memorization regimes; however, the corresponding results may depend on the specific properties of these sequences.

For document-level MIAs, utilizing a randomized split of ArXiv documents from The Pile along with the relevant models~\cite{biderman2023pythia,gpt-neo} offers a robust evaluation framework. Additionally, sampling ArXiv documents close to the training data cut-off date for an open-source model (RDD) might provide a valuable supplementary setup.

When resources allow, and if the motivation of developing MIAs is not strictly tied to pretraining, finetuning on randomized sets of sequences or documents would also be recommended to compare new MIAs. 

Finally, we encourage future research to always include a model-less baseline and to remain mindful of the inherent challenges (e.g. overlap, scaling from finetuning to pretraining) before interpreting results. 

Further information and resources to implement these benchmarks, alongside all the MIAs considered in this work, can be found in the Github repository we make available.


\bibliographystyle{plain}
\bibliography{bibliography.bib}

\onecolumn
\newpage
\appendix

\setcounter{section}{0} 
\renewcommand{\thesection}{A\arabic{section}} 

\section{Detailed MIA implementation}

\textbf{A. Detailed MIA implementation}

\vspace{1em}

Table~\ref{tab:mia_details} elaborates on the implementation details used for the MIAs from the literature we consider. Note that some MIAs, i.e. \textit{Loss}~\cite{yeom2018privacy}, \textit{Zlib}~\cite{carlini2021extracting} and \textit{Lower}~\cite{carlini2021extracting}, do not depend on the choice of certain hyperparameters. 

\begin{table*}[ht]
    \centering
    \caption{MIA Implementation Details Summary}
    \begin{tabular}{C{3cm}| p{12cm}}
    \toprule
        MIA & Hyperparameters and implementation details. \\
        \midrule
        \midrule 
        \textit{Ratio-LLaMA-2}~\cite{carlini2021extracting}  & We use LLaMA-2 7B~\cite{touvron2023llama2} as a reference model.\\
        \midrule 
        \textit{Neighborhood}~\cite{mattern2023membership}  & We follow the neighborhood generation procedure outlined in Section 2.2 of Mattern et al.~\cite{mattern2023membership} with RoBERTa~\cite{liu2019roberta} as an MLM, neighborhood size of $50$ samples, dropout probability $0.7$ and top-k = $10$.\\
        \midrule 
        \textit{Min-K\% Prob}~\cite{shi2023detecting}  & Following the best performing setting reported in Section 4.3 of Shi et al.~\cite{shi2023detecting}, we use $k = 20$. \\
        \midrule 
        \textit{Min-K\%++}~\cite{zhang2024min}  & In line with Min-K\% Prob, we use $k=20$. \\
        \midrule 
        \textit{ReCaLL}~\cite{xie2024recall}  & We use a fixed prefix with $10$ shots drawn from an auxiliary non-member distribution.\\
        \midrule 
        \textit{Probe (real)}~\cite{liu2024probing} & We use the code provided alongside Liu et al.~\cite{liu2024probing}. For proxy model training we use $4,000$ samples from the auxiliary non-member distribution, randomly split into equally-sized subsets of members and non-members. We then fine-tune our target model on the member subset for $10$ epochs with AdamW optimizer, constant learning rate of \num{1e-3}, default weight decay of \num{1e-2} and a batch-size of 100. We then train a logistic regression classifier for the activation vector, for 1000 epochs with AdamW optimizer, the constant learning rate \num{1e-3} and a weight decay \num{1e-1}.\\
        \midrule 
        \textit{PAC}~\cite{ye2024data}  & Using the implementation provided by Ye et al.~\cite{ye2024data}, we generate $10$ augmentations with alpha parameter of $0.3$. We the compute polarized distance using $k_\text{max}=0.05$ and $k_\text{min}=0.3$ following the best performing results reported in the original paper.\\
        \midrule 
        \textit{SURP}~\cite{zhang2024adaptive}  & Following the best performing setup reported in Section 5.3 of Zhang et al.~\cite{zhang2024adaptive}, we use $k=40$ with an entropy threshold of $2.0$\\
        \midrule 
        \textit{CON-ReCall}~\cite{wang2024recall}  & We use $10$-shot prefixes for both member- and non-member-conditioned likelihoods, drawn randomly and independently for each target example. Non-member prefixes are drawn from the auxiliary non-member distribution. For member prefixes we assume partial access by the attacker, and draw $10$ samples from the training dataset excluding the target example. \\
        \midrule 
        \textit{SaMIA}~\cite{kaneko2024sampling}  & Using the default setup in the implementation provided alongside Kaneko et al.~\cite{kaneko2024sampling}, we generate $10$ candidates using top-k sampling with top-k=$50$, temperature of $1.0$, and maximum length of $1,024$. We take the prefix to be $50$\% of the input sample.\\
    \bottomrule 
    \end{tabular}
\label{tab:mia_details}
\end{table*}

Note that we do not consider the MIAs requiring additional non-member data, i.e. \textit{ReCaLL}~\cite{xie2024recall}, \textit{Probe (real)}~\cite{liu2024probing} and \textit{CON-ReCall}~\cite{wang2024recall}, for the results in Table~\ref{tab:pile_auc} as we use all non-member data for evaluation. Generating additional, synthetic non-member data for the trap sequences in Table~\ref{tab:traps_auc} was feasible by following code made available by Meeus et al.~\cite{meeuscopyright}. 

Finally, for the results in tables~\ref{tab:traps_auc} and~\ref{tab:pile_auc}, we compute the AUC on $1,000$ bootstrapped subsets of members and non-members and report both the mean and standard deviation across all results~\cite{bertail2008bootstrapping}. 

\vspace{1em}

\textbf{B. Results for document-level MIAs}

\vspace{1em}

\textbf{Data collection.} We collect two datasets to evaluate document-level MIAs on LLMs: 

\begin{enumerate}
    \item ArXiv papers from the Pile. We randomly sample $1,000$ documents from the train set (members) and $1,000$ documents from the test set (non-members) from (the uncopyrighted version of) the Pile~\cite{pile_uncopyrighted}, ensuring that the selected documents have at least $5,000$ words (any sequences of characters seperated by a white space). We then split each document in $25$ sequences of $200$ words, and consider as full documents the first $5,000$ words. 
    \item ArXiv (RDD). We instantiate the RDD setup for the suite of OpenLLaMA~\cite{openlm2023openllama} models, for which the entire training dataset is known. As members, we consider papers from a month before the cutoff date for the training dataset of the target LLM  i.e. $13,155$ papers published in February 2023. We collect this set of documents using Github repository of RedPajama-Data~\cite{together2023redpajama}. For non-members, we consider the $16,213$ papers released in March 2023 - so one month after the member documents. We collect this set of documents through the Amazon S3 bucket hosted by ArXiv maintainers~\footnote{\url{https://info.arxiv.org/help/bulk_data_s3.html}}. We also apply the same preprocessing, and only consider documents of at least $5,000$ words, truncated to $5,000$ words and split into $25$ sequences of $200$ words. 
\end{enumerate}

\textbf{Models.} For the ArXiv papers from the Pile we use Pythia-dedup-6.9B~\cite{biderman2023pythia} as target model, while for the RDD setup we use OpenLLaMA-7B~\cite{openlm2023openllama}.

\textbf{MIA methodology.} We instantiate sequence- and document-level MIAs from the literature. We report both the AUC achieved for all sequences ($50,000$ in total) as well as the AUC when the sequence-level membership scores are scaled up to the document-level. For this, we consider the method proposed by Shi et al.~\cite{shi2023detecting} and described in Sec.~\ref{sec:unit_of_interest}. Namely, a membership inference signal is computed on the $k=25$ sequences $s_{tj}$ of $200$ words per document $d_t$. Next, an optimal threshold $\tau_{\text{seq}}$ is determined for the inference signal to maximize accuracy on the sequence-level MIA. By classifying the selected sequences using the optimal threshold, the average binary prediction is then computed for a document, and is used as document-level scoring function $y_{\text{doc}}(\mathcal{M}, d_t)$, or $y_{\text{doc}}(\mathcal{M}, d_t) = \frac{1}{k} \sum_{j=1}^k \hat{y}_{\text{seq}}(\mathcal{M}, s_{tj})$. Again, we do not consider the MIAs requiring additional non-member data, i.e. \textit{ReCaLL}~\cite{xie2024recall}, \textit{Probe (real)}~\cite{liu2024probing} and \textit{CON-ReCall}~\cite{wang2024recall}, and leave out \textit{SaMIA}~\cite{kaneko2024sampling} and \textit{Neighborhood}~\cite{mattern2023membership} due to the large computational costs required for this number of sequences.

We also instantiate the document-level MIA as proposed by Meeus et al.~\cite{meeus2024did}. We query the target LLM on sequences of length $2,048$ tokens and consider both their ratio and maximum normalization strategy using token frequency (\textit{RatioNormTF} and \textit{MaxNormTF}), across two feature aggregations (simple aggregate statistics \textit{AggFE} and histogram (\textit{HistFE})). 

In all document-level MIAs, we consider $1,600$ documents for training (i.e. determining the optimal threshold $\tau_{\text{seq}}$ in the method by Shi et al.~\cite{shi2023detecting}, and to train the meta-classifier for the method by Meeus et al.~\cite{meeus2024did}) and use the remaining $400$ documents for evaluation (averaged over $5$ splits). For the sequence-level MIAs, as above, we compute the AUC on $1,000$ bootstrapped subsets of members and non-members and report both the mean and standard deviation across all results~\cite{bertail2008bootstrapping}. 

\begin{table*}[ht]
    \centering
    \caption{Benchmarking sequence and document-level MIAs. Left: MIA AUC for ArXiv papers ($25$ sequences of $200$ words for both $1,000$ members and non-members) randomly sampled from train and test split from The Pile~\cite{pile_uncopyrighted}, attacking Pythia-dedup-6.9B~\cite{biderman2023pythia}. Right: MIA AUC for ArXiv papers (same length) randomly sampled from the RDD setup, attacking OpenLLaMA-7B~\cite{openlm2023openllama}.}
        \begin{tabular}{ccc|cc}
    \toprule
          & \multicolumn{2}{c}{ArXiv (The Pile~\cite{pile})} & \multicolumn{2}{c}{ArXiv (RDD)} \\
         MIA & Sequence & Document & Sequence & Document \\
         \midrule
         \midrule
         \textit{Loss}~\cite{yeom2018privacy} & $0.522\pm.003$ & $0.527\pm.034$ & $0.497\pm .002$ & $0.512\pm.020$ \\ 
         \cmidrule{1-5}
        \textit{Zlib}~\cite{carlini2021extracting} & $0.527\pm.003$ & $0.541\pm.028$ & $0.501\pm.003$ & $0.499 \pm .022$ \\ 
         \cmidrule{1-5} 
        \textit{Lower}~\cite{carlini2021extracting}& $0.514\pm.003$ & $0.522 \pm .028$ & $0.508 \pm .003$ & $0.507 \pm .024$ \\ 
         \cmidrule{1-5}
        \textit{Ratio-LLaMA-2}~\cite{carlini2021extracting}
        & $0.537\pm .003$ & $0.554\pm .030$ & $0.516\pm .003$ & $0.517\pm .003$ \\ 
        \cmidrule{1-5}
        \textit{Min-K\% Prob}~\cite{shi2023detecting} & $0.524\pm.003$ & $0.525\pm.031$ & $0.501 \pm .003$ & $0.496 \pm .009$ \\ 
        \cmidrule{1-5}
        \textit{Min-K\%++}~\cite{zhang2024min} & $0.519 \pm .003$ & $0.520 \pm .040$ & $0.499 \pm .003$ & $0.494 \pm .024$ \\ 
        \cmidrule{1-5}
        \textit{SURP}~\cite{zhang2024adaptive} & $0.521\pm.003$ & $0.526\pm.038$ & $0.500 \pm .003$ & $0.487 \pm .025$ \\
        \cmidrule{1-5}
        \textit{AggFE}+\textit{RatioNormTF}~\cite{meeus2024did} & \ding{55} & $0.548 \pm .017$ & \ding{55}& $0.508 \pm .046$ \\ 
        \cmidrule{1-5}
        \textit{AggFE}+\textit{MaxNormTF}~\cite{meeus2024did} & \ding{55} & $0.485 \pm .013$ & \ding{55}& $0.522\pm.055$ \\ 
        \cmidrule{1-5}
        \textit{HistFE}+\textit{RatioNormTF}~\cite{meeus2024did} & \ding{55} & $0.531 \pm .009$ & \ding{55}& $0.515 \pm .026$ \\ 
        \cmidrule{1-5}
        \textit{HistFE}+\textit{MaxNormTF}~\cite{meeus2024did} & \ding{55} & $0.513 \pm .026$ & \ding{55}& $0.513 \pm .040$ \\ 
        \midrule
         \bottomrule \\
    \end{tabular}
\label{tab:doc_level}
\end{table*}
\label{app:doc_level}

\end{document}